\documentclass{article} 
\usepackage[preprint]{iclr2024_conference}

\usepackage[utf8]{inputenc} 
\usepackage[T1]{fontenc}    
\usepackage{hyperref}       
\usepackage{url}            
\usepackage{booktabs}       
\usepackage{amsfonts}       
\usepackage{nicefrac}       
\usepackage{microtype}      
\usepackage{xcolor}         

\usepackage{algorithm}
\usepackage{algpseudocode}

\newtheorem{theorem}{Theorem}[section]
\newtheorem{defi}[theorem]{Definition}

\usepackage{booktabs}
\usepackage{multirow}
\usepackage{siunitx}
\usepackage{amssymb}
\usepackage{booktabs}       
\usepackage{amsfonts}       
\usepackage{nicefrac}       
\usepackage{microtype}      
\usepackage{amsmath}
\usepackage{graphicx}
\usepackage{multirow}
\usepackage{booktabs}
\usepackage{wrapfig}

\title{Large-Language-Model Empowered \\ Dose Volume Histogram Prediction for \\ Intensity Modulated Radiotherapy }

\author{%
  Zehao Dong\textsuperscript{1}  Yixin Chen\textsuperscript{1}  Hiram Gay\textsuperscript{2}  Yao Hao\textsuperscript{2} \\ \textbf{Geoffrey D. Hugo\textsuperscript{2} Pamela Samson\textsuperscript{2} Tianyu Zhao\textsuperscript{2} } \\
  \textsuperscript{1} Department of Computer Science $\&$ Engineering, Washington University in St. Louis \\
  \textsuperscript{2} Department of Radiation Oncology, Washington University School of Medicine \\
  \texttt{\{zehao.dong,hiramgay, yaohao,gdhugo,psamson,tzhao\}@wustl.edu} \\
  \texttt{chen@cse.wustl.edu} 
}
\begin{document}

\maketitle

\begin{abstract}
   Treatment planning is currently a patient specific, time-consuming, and resource demanding task in radiotherapy. Dose-volume histogram (DVH) prediction plays a critical role in automating this process.  The geometric relationship between DVHs in radiotherapy plans and organs-at-risk (OAR) and planning target volume (PTV) has been well established. This study explores the potential of deep learning models for predicting DVHs using images and subsequent human intervention facilitated by a large-language model (LLM) to enhance the planning quality. We propose a pipeline to convert unstructured images to a structured graph consisting of image-patch nodes and dose nodes. A novel Dose Graph Neural Network (DoseGNN) model is developed for predicting DVHs from the structured graph. The proposed DoseGNN is enhanced with the LLM to encode massive knowledge from prescriptions and interactive instructions from clinicians. In this study, we introduced an online human-AI collaboration (OHAC) system as a practical implementation of the concept proposed for the automation of intensity-modulated radiotherapy (IMRT) planning. In comparison to the widely-employed DL models used in radiotherapy, DoseGNN achieved mean square errors that were 80$\%$, 76$\%$ and 41.0$\%$ of those predicted by Swin U-Net Transformer, 3D U-Net CNN and vanilla MLP, respectively. Moreover, the LLM-empowered DoseGNN model facilitates seamless adjustment to treatment plans through interaction with clinicians using natural language. 
   
\end{abstract}
\section{Introduction}
\label{intro}
Radiotherapy is a crucial component in cancer treatment. Technological innovations ~\cite{jaffray2015radiation} have driven the transition from conformal radiation therapy to intensity-modulated radiation therapy (IMRT) for sufficient planning target volume (PTV) coverage and organ-at-risks (OARs) sparing. Nowadays, IMRT ~\cite{palta2003intensity,bortfeld2006imrt} has been the primary delivery technique in radiotherapy. However, the process of creating a desirable treatment plan is typically time-consuming and expensive, and the plan quality heavily relies on the experience and skills of individual planners and institutions. Thus, automatic treatment planning has received a growing attention as a promising alternative addressing the need for fast, effective, and customized IMRT planning. The primary goal of IMRT treatment planning is to produce a cumulative dose distribution that satisfies both the dose prescribed to tumors while meeting the normal tissue dose constraints. Consequently, Dose-volume histogram (DVH) prediction ~\cite{appenzoller2012predicting} has been widely used as quantitative metrics for evaluating  plan quality and played a critical role in automatic treatment planning. Numerous efforts ~\cite{zarepisheh2014dvh} are put into generating DVH-based optimization algorithms for radiotherapy automation and re-planning.

To provide efficient treatment planning pipeline, knowledge-based planning (KBP) methods ~\cite{cagni2017knowledge, wu2009patient, wang2015patient} and deep learning (DL) methods ~\cite{sumida2020convolution,xing2020feasibility, campbell2017neural} are developed. KBP methods have been widely investigated and clinically implemented to utilize a database of prior treatment plans for leveraging the planning expertise of physicians and physicists. Knowledge-based planning assistant tools like RapidPlan have widely applied in clinics and researches. On the other hand, DL methods take advantage of the advancement in artificial intelligence. Various advanced deep learning models, such as 3D U-net ~\cite{cciccek20163d} and Swin Transformer ~\cite{liu2021swin}, are applied to discover intricate structures in high-dimensional CTs and to perform the end-to-end training from the medical raw data. DL methods have achieved state-of-the-art performance in many medical applications ~\cite{momin2021knowledge} and received increased interest in the last decade.  

With the advancement of radiological science and computer science, CT images have been widely been used in guiding the delivery of radiation dose  ~\cite{fan2020data,liu2019deep,dong2024dosegnn, kearney2018dosenet}. Typically, convolutional neural networks, such as U-net \cite{u-net}, are utilized to extract useful information from CT images of helical tomotherapy for the prediction of volumetric dose distribution. However, the feasibility and performance of DL methods predicting DVHs on uncorrected CT 3D images haven’t been thoroughly studied and understood \cite{dong2023performance}. In this problem, 3D CT images and relevant meta clinical data are taken as inputs, then the objective is to predict the corresponding 3D dose matrices that specifies the radiation dose values to deliver. Three properties make the deep learning problem difficult: (1) As a geometric object, the learned features of each point in the 3D CT images should be invariant to certain transformations. For example, translating points all together should not modify the prescribed dose values related to them. (2) The shapes/resolutions of the input 3D CT images and the output 3D dose value matrix can be different across samples, which requires the deep learning model to be capable of manipulating input of different shapes and generating predictions of sample-dependent shapes. (3) CT images and dose matrices in the DVH prediction task are usually large, causing heavy computation burden and complexity.

To address the challenge, we introduce a pipeline to convert unstructured images to structured data resistant to translations, perturbations and noises in 3D CT images, based on which deep learning models are implemented for the dose prediction. The proposed image-conversion pipeline resorts to the the positional encoding scheme widely applied in sequence encoding~\citep{vaswani2017attention} and graph representation learning ~\citep{yan2020does, dong2022pace}. Since DVHs in radiotherapy plans show close relations ~\cite{appenzoller2012predicting,wang2020review} to the geometric information of organ-at-risk (OAR) and planning target volume (PTV), each pixel in the image is associated with a positional encoding representing the positions of PTV and OARs as well as their relative geometric relations.  Specifically, image segmentation algorithms ~\cite{haralick1985image,minaee2021image} are proven to efficiently extract contours of OARs from uncorrected images, while PTV surfaces are provided in prescriptions from DICOM ~\cite{mildenberger2002introduction} documentation of treatment plans. Then the one-hot features of the OARs/PTV indicators and geometry features (Manhattan distance, angle) to the center of PTV are used in the positional encoding scheme.  

To effectively extract useful information from the structured data after conversion and solve the mismatch between the dose matrix and CBCT image, we develop a novel DL model, DoseGNN, to combine the strength of vision Transformers ~\cite{carion2020end,dosovitskiy2020image} in image feature extraction and the power of GNNs~\cite{kipf2016semi, Hamilton2017,xu2018how, zhang2018end, zhang2021nested} in encoding graph-structured relational data. Since dose (3D dose matrix) and image have different shapes and resolutions, we propose to modulate their relations by formulating a dose-image graph. For each grid/pixel correlation in the dose / image pair, its’ geometric position is available from the DICOM data, then the grid/pixel correlation is associated with a dose/image pair that takes the resolution as the edge length and is centered at the grid/pixel’s geometric position. Then, in the formulated graph, each node represents a dose or an image voxel, while an edge connects a dose node and an image node if the overlap ratio of the corresponding dose and image voxel is above a pre-defined threshold, like $0.3$ in common cases. In the formulated dose-image graph, the features of image nodes are learnt through a hierarchical vision Transformer  (i.e. Swin Transformer)~\cite{liu2021swin},while dose node takes the binary encoding of the OARs/PTV indicator as the initial node feature. Then, a GNN model is used to propagate information in the dose-image pair to extract the vector representations of each dose node for predicting its’ dose value. 

In addition, The challenging nature of data-driven DVH prediction and treatment planning is often further complicated by a lack of high-quality treatment plans for training, then the trained deep learning models can not well characterize all potential patient's medical condition and anatomy in practice. To alleviate this challenge, an intuitive option is to enable the DL models to incorporate the prescriptions full of medical records and the instructions of doctors/experts with domain knowledge to correct the outlier predictions.  Inspired by the recent advancements of large language models (LLMs), such as OpenAI's ChatGPT ~\cite{koubaa2023gpt,brown2020language} and Google's PaLM ~\cite{chowdhery2022palm}, we take embedding-visible LLMs ~\cite{chen2023exploring, he2020deberta} as the enhancers that utilize the powerful text understanding ability and remarkable zero-shot capability of LLMs to improve the DL model performance. In the proposed DoseGNN model, we introduce a prompt node in the dose-CBCT graph, which is connected with dose nodes in the formulated graph. The prompt node takes prescriptions and doctors' instructions as text input, then embedding-visible LLMs like Sentence-BERT~\cite{reimers2019sentence} and Deberta~\cite{he2020deberta} are implemented to enhance text attributes directly by encoding them as initial node features of the prompt node.

\begin{figure}[t]
\begin{center}
\centerline{\includegraphics[width=0.9\textwidth]{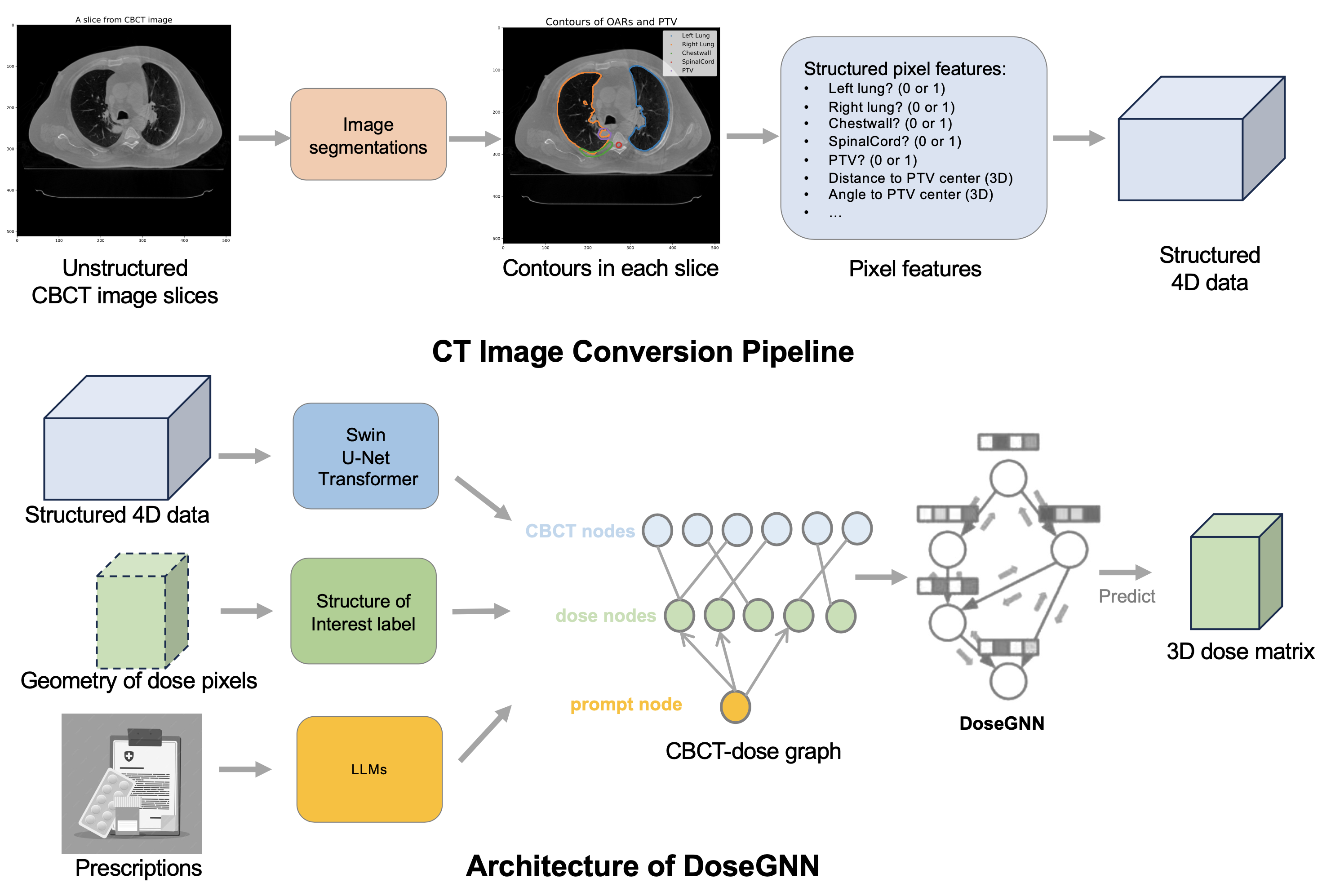}}
\caption{. Illustration of image conversion pipeline and DoseGNN model. The image conversion pipeline converts unstructured 3D images to structured 4D tensor/data that summarizes the contour information and geometric relation of OARs and PTVs. In DoseGNN, Image processing deep learning models (like swin Transform) and LLMs are used to convert the structured 4D tensor and prescriptions/instructions text to node features in an image-dose graph, then a GNN is applied to predict dose values of dose nodes in the graph.}
\label{fig:fig1}
\vskip -0.15in
\end{center}
\end{figure}

We evaluate the proposed image conversion pipeline and LLM enhanced DoseGNN model in one of the most frequently investigated disease sites: lung cancer. Comprehensive experiments demonstrate the advantage of our DoseGNN model over existing deep learning models capable of making predictions from images. Across all structure of interests (OARs and PTVs), DoseGNN achieves the significant lower error in predicting the max dose as well as the mean dose. In addition, comparison of DVHs of baseline doses in plans created by experienced dosimetrists and predicted doses from different deep learning models indicates that DoseGNN has the state-of-the-art capability to approximate the prescribed dose distribution, which facilitates downstream tasks, such as assessing the plan quality, identification of sub-optimal plans, and knowledge-based automatic planning. 


\section{Methods}
\label{metho}
In this section, we introduce the image-conversion pipeline and the proposed DoseGNN.  Figure ~\ref{fig:fig1} illustrates the overall architecture.

\subsection{Problem Formulation}

At first, we introduce basic concepts and notations in the DVH prediction problem. Basically, the studied DVH prediction problem takes as inputs the image data and medical text data, and then the objective is to predict the corresponding dose image/matrix. 

\paragraph{Image and dose data} Image and dose data are tuples $(X, s_c, r_c, z_c)$ and $(Y, s_d, r_d, z_d)$.
(1) $X$ is the 3D image tensor and $Y$ is the 3D dose value tensor to predict. The sizes of tensor $X/Y$ illustrates the sizes of the 3D medical CT images/dose image for prediction, while each entry of tensor X/Y associates with a geometric position of medical image/dose image. (2) In general, medical imaging processing, the spacings in axis $x$, $y$, $z$ are not necessarily isotropic. In the studied problem, we observe that the spacings in axis $x$ and $y$ for each slice of CT/dose image are same, then we term this shared spacing in each slice as the slice resolution, and use $r_c$  and $r_d$ to denote the slice resolution of CT data and dose data, respectively. Then, to fully preserve the geometric information of CT images and dose images, we use 2-dimensional vectors  $s_c$ and $s_d$ to indicate the origin of each slice, which are respectively the geometry position of the top left pixel/entry of each slice of CT images and dose images. In the end, we use vectors $z_c$ and $z_d$ to store the z values of geometric position of each slice of CT images and dose images.

Images $X$ are a fundamental three-dimensional description of patients’ OARs and PTVs. CT images are particularly useful in radiation oncology for treatment planning and localization of tumors, as they provide detailed information about the size, shape, and location of the tumor and surrounding tissues for delineation, in addition to electron density for dose calculation. Consequently, images X are used as inputs to deep learning models to precisely predict the 3D dose image/matrix Y that targets the radiation beams at the tumor, minimizing exposure to healthy tissues and improving the improving the coverage of tumor.

Since images and dose do not always have the same resolution, more geometric information is required to reveal the relation of voxels in the image tensor $X$ and dose tensor $Y$. Each entry in the image tensor $X$ or dose image tensor $Y$ represents a voxel, a specific volume of tissue in the patients' body. Then, in the image tensor $X$, entry $(x,y,z)$ corresponds to a voxel coordinated (the 3D coordinate is $(\hat{x}, \hat{y}, \hat{z})$) by $s_c[0] + (x-1) \times r_c \leq \hat{x} \leq s_c[0] + x \times r_c$, $s_c[1] + (y-1) \times r_c \leq \hat{y} \leq s_c[1] + y \times r_c$, $z_c[z-1] \leq \hat{z} \leq z_c[z]$. Similarly, in the dose image tensor $Y$, entry $(x,y,z)$ corresponds to a voxel coordinated by $s_d[0] + (x-1) \times r_d \leq \hat{x} \leq s_d[0] + x \times r_d$, $s_d[1] + (y-1) \times r_d \leq \hat{y} \leq s_d[1] + y \times r_d$, $z_d[z-1] \leq \hat{z} \leq z_d[z]$.

\paragraph{Medical text data} Medical text data T are prompt-oriented text documents of clinician’s instructions and prescriptions to describe the goals of the treatment as well as the medical factors of a patient. The radiotherapy treatment planning is widely depended on various factors, including the type of cancer, the location of the tumor, the size of the tumor, the overall health of the patient, the goals of the treatment, etc. Overall, the configurations of tumors and other health organs at risk are provided in the image data. In this paper, we propose to formulate other medical record data as medical text data T, which can be used as prompts to a large language model to promote the performance of deep learning model. 

\subsection{Image Conversion Pipeline}

In the modern medical and oncology research, unstructured data are almost everywhere, including CBCT images, prescription text files, etc. The recent advancement of GPT has revolutionized the field of machine learning on texts, providing the general solution to learn from contextualized medical text file. However, it is hard to effectively analyze CT images in the DVH prediction task. Various deep learning models have been proposed for machine learning on 3D data. Volumetric CNNs (V-CNNs such as VoxNet \cite{maturana2015voxnet} and ShapeNet \cite{wu20153d}) and their variants \cite{li2016fpnn, wang2015voting} developed 3D convolution operations on voxelized shapes and are widely applied to volumetric segmentation and classification of 3D data like medical imaging (computed tomography, magnetic resonance imaging, etc.). However, V-CNNs usually suffer the out-of-memory problem in DVH prediction due to the extreme large size  of CT images, and the volumetric representation of 3D CT images also causes limited learning ability. On the other hand, multiview CNNs \cite{su2015multi,qi2016volumetric} (render 3D shapes into 2D images and then apply 2D convolution networks. However, these methods did not support pixel/point level prediction tasks. Overall, dominant deep learning approaches for 3D images are not suitable in DVH prediction task.

To address this challenge, we resort to feature-based methods \cite{guo20153d,fang20153d} in 3D data representation learning. The close correlation between the dose distribution and the geometry positions of PTV and OARs has been well studied. Hence, we propose the image conversion pipeline to convert the 3D images to structured 4D tensor enriched with geometric information of OARs and PTVs of the input 3D images, reserving informed geometry information that enables accurate diagnoses and treatment planning. 

Overall, the image conversion pipeline consists of two steps: 1) image segmentation, and 2) pixel feature extraction. In the image segmentation process, each slice of the input 3D image is processed through the image segmentation algorithms to extract the contour of each OAR and PTV. After that, the pixel feature extraction step extracts a feature vector for each pixel in the input 3D image to characterize its geometry information relevant to the position of OARs/PTV as well as their relations. Specifically, we use the binary $0/1$ indicator to show whether a pixel/entry is within the contour of an OAR/PTV, then we use the distance and the angle to the center of the PTV to characterize the geometric relation of the PTV and nearby organs/tissues that are sensitive to the effects of radiation therapy. Specifically, $14$ OARs are studied in the paper: body, left and right brachial plexus, esophagus, skin rind, bronchial tree, chest wall, left and right lung, heart, trachea, carina, spinal cord and liver. In the case that the patient was missing one of these OARs, the corresponding channel was set to 0 for the input. Consequently, the 3D images are transformed as a structured 4D tensor with geometry information of OARs and PTV.

\subsection{DoseGNN Model}

Next, we introduce the proposed deep learning model, DoseGNN, for precise prediction of dose image/matrix in the DVH prediction task. Two challenges make the prediction/regression task indirect and difficult. First, for each input image data, the shape of the image tensor X and the 3D dose tensor Y to predict are different, which requires the deep learning model to embed the relation between voxels in image and the dose. Second, the shapes of the image tensors and the dose tensors across patient samples are different, indicating that a proper deep learning model must be able to handle inputs of different sizes and generate outputs for arbitrary number of voxels. To address these challenges, we propose to formulate an image-dose graph to embed the image-dose voxel relation, and then use a heterogeneous GNN to predict the dose values assigned to dose voxels by propagating message in the image-dose graph. 

\begin{defi}
    (Image-dose graph) Image-dose graph is a heterogeneous graph  $\mathcal{G} = (\mathcal{G}, \mathcal{E})$, in which $\mathcal{V}$ and $\mathcal{E}$ represent the node set and the edge set, respectively. A mapping function $\phi: \mathcal{V} \to \mathcal{A}$ assigns each node $v \in \mathcal{V}$ a node type $\phi(v)$, where $\mathcal{A} = \{\textit{n}_{ct}, \textit{n}_{d}, \textit{n}_{p} \}$ is the node type set. Here $\textit{n}_{ct}$, $\textit{n}_{d}$, $\textit{n}_{p}$ denote the nodes associated with image voxels, dose voxels and medical text prompts, respectively.
\end{defi}

\paragraph{Node attributes/features} In the heterogeneous image-dose graph, the attributes/features of different types of nodes and edges have different meanings. For the notation simplicity, we use the $f_{img}$ and $f_{llm}$ to denote the image processing deep learning models and the large language models, respectively. 

\begin{itemize}
    \item (Image nodes) Each image node $v$ (i.e. $\phi(v) = \textit{n}_{ct}$) represents a voxel in the input image and can be indexed by it's coordinates $(x_1,y_1,z_1)$. Let $X_{CT}$ be the structured 4D tensor generated by the proposed image conversion pipeline, DoseGNN proposes to extract useful geometric information with a image processing deep learning model $f_{img}$, then the initial node features of the CBCT node $v$ is $h^{0}_{v} = f_{img}(X_{CT})[x_1,y_1,z_1]$. 
    \item (Dose nodes) Each dose node $v$ (i.e. $\phi(v) = \textit{n}_{d}$) represents a voxel in the dose image and can be indexed by it's coordinates $(x_2,y_2,z_2)$. Consequently, the dose node $v$ is associated with a dose value $Y[x_2,y_2,z_2]$ to predict. The dose voxel can overlap many voxels in the input image, and each image is assigned with an indicator of the OARs. Then, the dose node $v$ takes the prescription dose and the one-hot embeddings of the indicator of OAR with the maximal voxel coverage as the initial node features $h^{0}_{v}$.
    \item (Prompt node) The prompt node $v$ (i.e. $\phi(v) = \textit{n}_{p}$) represents an input medical text data $X_{MT}$ such as the formulated prescriptions and the doctors' instructions. Then we take the output embedding of a fine-tuned LLM $h^{0}_{v} = f_{llm}(X_{MT})$ as the initial node features.     
\end{itemize}

In this pipeline, LLMs are leveraged to enhance the medical text attributes. LLMs has been proven to be fine-tuned to perform well in the chemical domain \cite{jablonka2023gpt}, where a string representation (the chemical formats like SMILES or SELFIES) is provided as a prompt to the LLMs. Following this inspiration, we first fine-tune the LLMs on the downstream dataset of medical texts. Subsequently, the text embeddings engendered by the fine-tuned LLMs are employed as the initial node features of the prompt node. The formulation of prompts and completion plays a critical role as it affects the ability of the LLMs to capture their relationship as well as the computational cost of fine-tuning the LLMs. 


\paragraph{Edge configuration} Image-dose graph embeds the connection of image voxels and dose voxels according to their geometry position. In the image-dose graph, image nodes and dose nodes represent voxels in images and dose, respectively. Then, an image node is connected with a dose node if and only if the proportion of the coverage of their voxels and the smaller voxel is larger than a threshold. On the other hand, the prompt node is connected with the dose nodes.

\paragraph{Homogeneous GNN} In the end, we utilize a homogeneous GNN $g$ to predict the dose values associated with dos nodes in the image-dose graph $\mathcal{G}$. Since we use different scheme/models to generate the initial features of image nodes, dose nodes and the prompt node, the heterogeneity is encoded in the initialized node features. Then, the DVH prediction task is equivalent to a general node regression problem, and the proposed DoseGNN resorts to the homogeneous GNN to achieve the prediction purpose. Mean squared error between the predicted dose and the clinically delivered dose was used as the loss function, which is formulated as following, 
\begin{align}
    \textit{Loss} = \frac{1}{ \sum^{N}_{i=1} |D_i|} \sum^{N}_{i=1}\sum_{v \in D_i} (g(\mathcal{G}_{i}, H^{0})[v] - Y[x_v, y_v, z_v])^2
\end{align}
where $N$ is the number of samples, $D_i = \{v | v \in \mathcal{G}_{i}, \phi(v) = \textit{n}_{d}\}$ is the set of dose nodes in the $i$-th sample, and $(x_v, y_v, z_v)$ is the coordinates in dose image of the dose node $v$.

In this study, the DoseGNN model was trained on $40$ IMRT lung treatment plans that were created on images acquired on $\textit{Varian}$ $\textit{HyperSight}^{TM}$, a high performance CBCT (Cone beam CT) imaging system cleared for simulation and dose calculation by FDA.

\section{Experiments}
\label{exper}
In this section, we evaluate the effectiveness of proposed DoseGNN model and its interaction with clinicians. To illustrate their priority, we conduct experiments on lung cancer IMRT plans to show that: 

\begin{itemize}
    \item \textbf{1:} DoseGNN can achieve the most accurate prediction of the clinically delivered dose and significantly outperforms existing state-of-the-art deep learning models.
    \item \textbf{2:} LLMs-based interaction can effectively enhance the performance of treatment planning algorithms by incorporating clinician’s instructions/prescriptions. 
\end{itemize}

\subsection{Datasets, Baselines and Experiment Configuration}
\label{subsec:datasets}

Our experiments are conducted on 40 lung cancer treatment plans. Due to the size of the dataset, we perform 5-fold cross validation for robust comparison. All experiments are implemented in the environment of PyTorch using NVIDIA A40 GPUs. The training protocols is composed of the selection of the evaluation rates and training stop rules. Specifically, the learning rate of optimizer picks the best from the set $\{1e-4, 1e-3, 1e-2\}$; the training process is stopped when the validation metric does not improve further under a patience of $10$ epochs.

The studied DVH prediction problem is essentially a image prediction problem at the pixel level. Consequently, dominant image processing AI models are ideal baselines. In the experiment, we select three powerful baseline models from recent advancements in the field of image machine learning: Swin Transformer ~\cite{liu2021swin}, U-Net (CNN) ~\cite{ronneberger2015u}, and a vanilla MLP.

\subsection{Comparison of Predictive Performance}

In the experiment, we numerically evaluate how well DoseGNN can predict the clinically delivered dose in IMRT plans. Three evaluation metrics are used: (1) the mean square error (MSE) of the predicted dose and the clinically delivered dose; (2) the absolute error of the structure max and mean doses ($D_{max}$ and $D_{mean}$). We use the MSE loss to quantify the overall predictive performance across all structures (i.e. OARs, PTV, etc). On the other hand, the structure-based absolute errors are normalized by the prescription dose: $\frac{\hat{D}_{max} - D_{max}}{\textit{prescription dose}}$ and $\frac{\hat{D}_{mean} - D_{mean}}{\textit{prescription dose}}$, where $\hat{D}_{max}$ and $\hat{D}_{mean}$ are computed from predicted doses, $D_{max}$ and $D_{mean}$ are based on true delivered doses. Here we select 5 structures of interests: PTV and 4 lung-cancer related OARs: left lung, right lung, chest wall, and spinal cord.

\begin{figure}[t]
\begin{center}
\centerline{\includegraphics[width=0.99\textwidth]{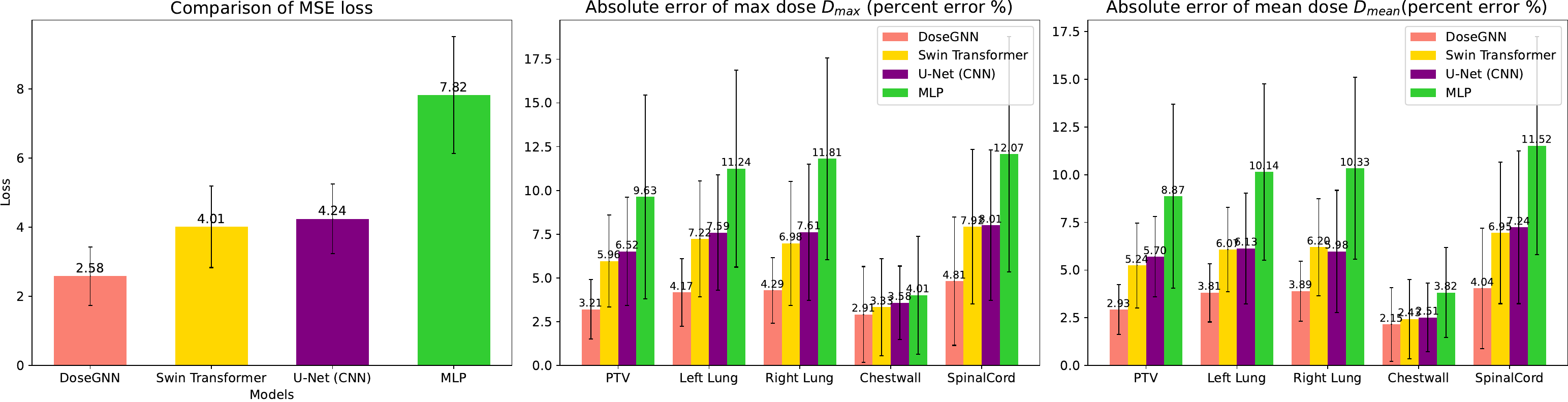}}
\caption{Comparison of predictive performance of deep learning models in DVH prediction.}
\label{fig:fig3}
\vskip -0.15in
\end{center}
\end{figure}

\begin{table}[t]
\centering
\resizebox{0.99\textwidth}{!}{
\begin{tabular}{lcccccccccccc}
\toprule
& \multicolumn{2}{c}{Overall $\downarrow$} & \multicolumn{5}{c}{Structure-based $D_{max}$ $\downarrow$} & \multicolumn{5}{c}{Structure-based $D_{mean}$ $\downarrow$}\\ 
\cmidrule(r){2-3}  \cmidrule(r){4-8} \cmidrule(r){9-13}
&  MSE Loss & & PTV & left lung & right lung & chest wall & spinal cord & PTV & left lung & right lung & chest wall & spinal cord \\
\midrule
MLP & 7.82 $\pm$ 1.69 & &
9.63 $\pm$ 5.84 & 11.24 $\pm$ 5.61 & 11.81 $\pm$ 5.75 & 4.01 $\pm$ 3.36 & 12.07 $\pm$ 6.73
& 8.87 $\pm$ 4.82 & 10.14 $\pm$ 4.62 & 10.33 $\pm$ 4.77 & 3.82 $\pm$ 2.36 & 11.52 $\pm$ 5.72\\

U-Net (CNN) & 4.24 $\pm$ 1.01 & &
6.52 $\pm$ 3.12 & 7.59 $\pm$ 3.29 & 7.61 $\pm$ 3.94 & 3.58 $\pm$ 2.08 & 8.01 $\pm$ 4.33
& 5.70 $\pm$ 2.09 &  6.13 $\pm$ 2.88 & 5.98 $\pm$ 3.21 & 2.51 $\pm$ 1.84 & 7.24 $\pm$ 4.02\\

Swin Transformer & 4.01 $\pm$ 1.18 & &
5.96 $\pm$ 2.63 & 7.22 $\pm$ 3.31 & 6.98 $\pm$ 3.54 & 3.33 $\pm$ 2.78 & 7.92 $\pm$ 4.41
& 5.24 $\pm$ 2.23 & 6.07 $\pm$ 2.21 & 6.20 $\pm$ 2.54 & 2.43 $\pm$ 2.08 & 6.95 $\pm$ 3.72\\

\midrule
\textbf{DoseGNN(no LLMs)} & 3.42 $\pm$ 0.87 & &
3.67 $\pm$ 1.73 & 4.39 $\pm$ 1.89 & 4.81 $\pm$ 1.88 & 3.07 $\pm$ 2.66 & 5.35 $\pm$ 3.62
& 3.29 $\pm$ 1.38 & 4.22 $\pm$ 1.50 & 4.15 $\pm$ 1.59 & 2.39 $\pm$ 1.93 & 4.89 $\pm$ 3.20\\

\textbf{DoseGNN} & \textbf{2.58} $\pm$ \textbf{0.93} & &
\textbf{3.21} $\pm$ \textbf{1.82} & \textbf{4.17} $\pm$ \textbf{1.92} & \textbf{4.29} $\pm$ \textbf{2.01} & \textbf{2.91} $\pm$ \textbf{2.74} & \textbf{4.31} $\pm$ \textbf{3.81}
& \textbf{2.93} $\pm$ \textbf{1.51} & \textbf{3.81} $\pm$ \textbf{1.61}& \textbf{3.89} $\pm$ \textbf{1.66}& \textbf{2.15} $\pm$ \textbf{2.05}& \textbf{4.04} $\pm$ \textbf{3.44}\\
\bottomrule
\end{tabular}
}
\vspace{5pt}
\caption{Evaluation of dose prediction ability. $5$-fold cross validation is performed. t-test is used to compare the mean MSE of different models, and the \textbf{best} results are \textbf{highlighted} if the mean MSE is significantly smaller ($p < 0.05$) than other methods. DoseGNN with LLMs achieves the lowest MSE loss and can provide most accurate prediction of delivered doses.}
\label{table:2}
\end{table}

Figure ~\ref{fig:fig3} and Table ~\ref{table:2} present the experimental results. We observe that the proposed DoseGNN can significantly achieve more accurate dose prediction in both overall evaluation and structure-based evaluation. The experimental results indicates that the CBCT conversion pipeline and CBCT-dose graph in DoseGNN model helps to better characterize the relation between the clinical dilivered doses and geometry information of OARs/PTV.

\begin{figure}[t]
\begin{center}
\centerline{\includegraphics[width=0.99\textwidth]{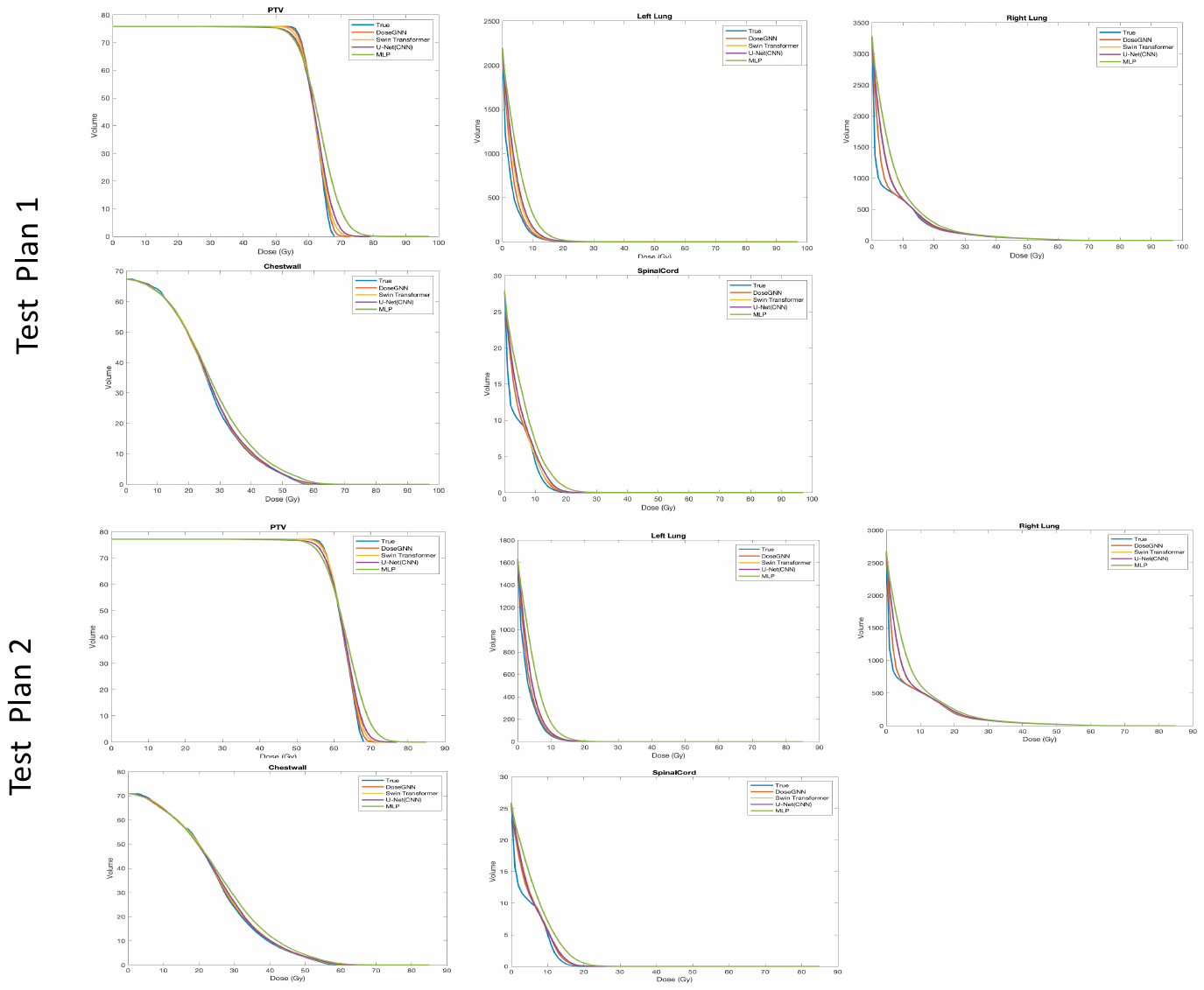}}
\caption{Comparison of predicted cumulative DVH (CDVH) against the CDVH of delivered dose}
\label{fig:fig4}
\vskip -0.15in
\end{center}
\end{figure}

\begin{figure}[t]
\begin{center}
\centerline{\includegraphics[width=0.66\textwidth]{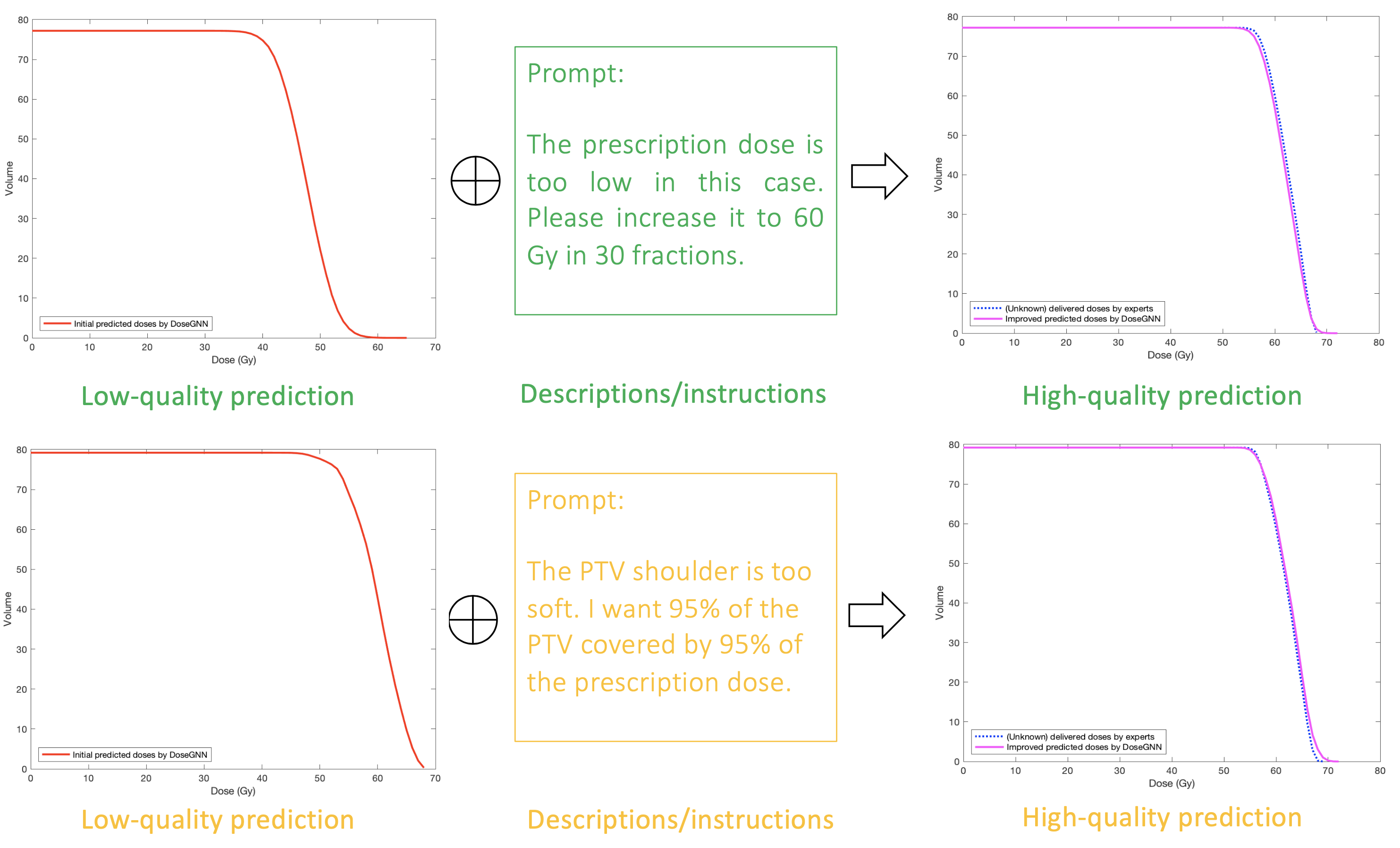}}
\caption{Clinicians can directly optimize the outcome of DoseGNN model with prompts/ instructions. In the figure, dashed blue lines indicate CDVHs computed from unknown delivered doses in clinical plans designed/optimized by human experts. A low-quality prediction of doses shows significantly different shape of CDVH and can be identified by doctors and other expert users without knowing the dashed blue lines.}
\label{fig:fig5}
\vskip -0.15in
\end{center}
\end{figure}

\subsection{Visualization of Cumulative DVH}

In the experiment, we visualize the cumulative dose-volume histograms (CDVH) to demonstrate the real-world impact of propsoed model.

Figure ~\ref{fig:fig4} compare the CDVHs of different deep learning models against CDVH of true clinical delivered dose
on five structure of interests (PTV and four OARs: left lung, right lung, chest wall, spinal cord). We find that DoseGNN has the best ability to capture the shape of ture CDVH among all the deep learning model. Since lots of knowledge-based automatic treatment planning or re-planning algorithms are DVH/CDVH-based, and a better prediction of DVH/CDVH can contribute to better treatment automation, our DoseGNN demonstrates the superiority than other models in real-world tasks.

\subsection{Effectiveness of LLMs}

In the end, we provide an example to show that further adjustment and personalization of the treatment plans can be achieved by adding additional layer of LLMs between clinicians and the DoseGNN model. Figure ~\ref{fig:fig5} illustrates two examples where predicted doses significantly fail to provide a DVH curve that describes the true plan. Based on the initial results, the experienced doctors can easily find that the predicted DVH fail to approximate a potential high-quality plan. Consequently, doctors can provide text instructions/prompts to LLMs for better dose predictions, and the instructions/prompts are then encoded by the fine-tuned LLMs as input to the DoseGNN to re-predict the delivered doses. The result shows that DoseGNN’s performance is significantly enhanced by the provided prompts (i.e. instructions/prescriptions). Furthermore, Table ~\ref{table:2} also quantitatively shows that LLMs help to improve the accuracy of predicted doses.

\section{Discussion}
\begin{figure}[t]
\begin{center}
\centerline{\includegraphics[width=0.99\textwidth]{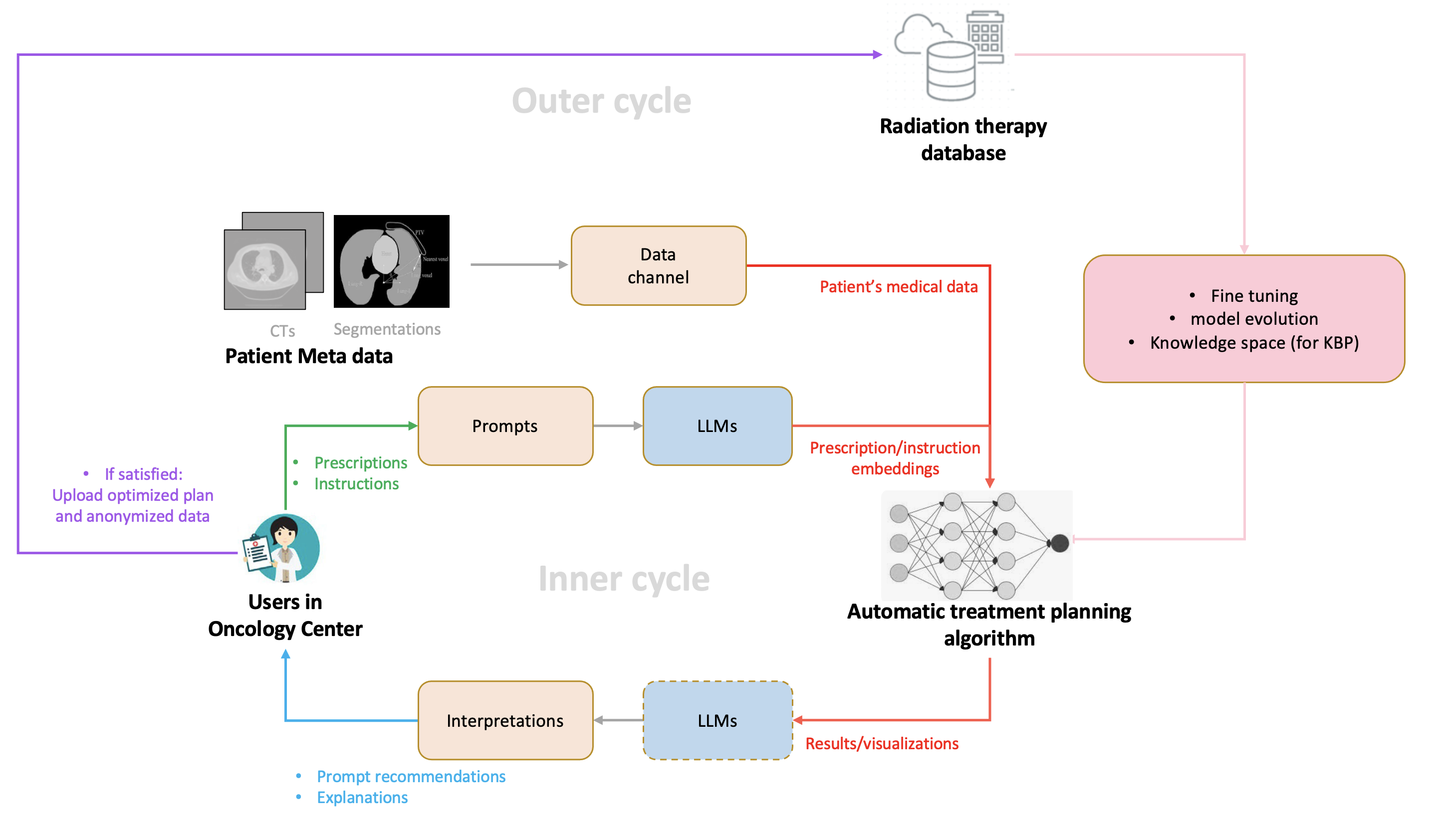}}
\caption{Illustration of the OHAC (online Human-AI collaboration) system.}
\label{fig:fig2}
\vskip -0.15in
\end{center}
\end{figure}

There are currently several limitations in this study and similar studies published regarding the data-driven deep learning models for automatic treatment planning. Our next focus is to address the most challenging one, which is the scarcity of high-quality treatment plans for training in a cost-effective manner. Typical deep learning algorithms use considerably larger datasets in the hundreds of thousands or millions if needed, which can be prohibitively expensive to acquire.  Therefore, the number of available IMRT plans for training in any specific deep-leaning model is usually limited, often representing the experience of a single-institution’s experience. This limitation significantly restricts the applicability and scalability of existing models. To overcome this hurdle, we propose to establish an online Human-AI collaboration (OHAC) system. This system will enable clinicians from cancer centers of varying in geographic locations and clinical settings to interact with the same deep learning models through an interface powered by LLM to facilitate the collection of evaluation and alternations(s) suggested by clinicians. The modified plans provided by clinicians will serve as high-quality input for continuous training and improvement of the deep learning model. 

For eventual clinical implementation, another key challenge is that deep learning models including those studied in this work generally predict a dose distribution rather than a set of machine instructions to deliver said dose distribution. Work is needed to generate deliverable IMRT plans (a sequence of machine instructions including collimator and gantry positions and monitor units to deliver) that can generate the predicted dose distributions.

The overall structure of the OHAC system is demonstrated in Figure ~\ref{fig:fig2}. The system consists of an inner patient-specific AI-based radiotherapy plan optimization cycle and an outer AI model evolution cycle based on an evolving radiation therapy database. Informally, the inner cycle, implemented in this study, integrates well-trained automation AI algorithms, clinician’s expertise, and LLMs to automatically optimize radiotherapy plans for each new patient in a radiotherapy center. The optimized plans will then be anonymized and uploaded to our radiation therapy database to support the fine-tuning and refinement of AI automation algorithms in the outer evolution cycle. Specifically, three user interface (UI) channels are used to provide configurations of cancer patients, optimization goals for automation algorithms on the platform, and the evaluation of optimized plans, separately. 1) The first channel (patient meta data channel) anonymizes and uploads the clinical metadata of a patient, such as CT images and other personalized medical information, for data characterization; 2) the second channel (prompt channel) allows clinicians to dictate their orders, including findings, treatment goals, specific areas of attention that should be incorporated into the plan; 3) the third channel (interpretation channel) shares evaluations and take feedback on the optimized treatment plans generated by the IMRT automation algorithm. LLMs are used in the second UI channel to explain the instructions/prescriptions from human experts and to communicate with automation algorithms on the platform. Based on user instructions/prescriptions from the second UI channel, the IMRT automation algorithms, such as the proposed DoseGNN model, take advantage of the superior power of LLMs in understanding textual information, so that these automation algorithms can be enhanced by enriched medical records and factors. Based on evaluations from the third UI channel, the treatment plans are optimized in an iterative way to provide high-quality plans based on direct dictation from clinicians. Furthermore, the interactive process between the platform and massive radiotherapy centers also provides abundant training samples for the platform to support the training of more sophisticated algorithms. In summary, the platform-user collaborative IMRT automation system enables a cycle of instruction, automated planning, and evaluation. Therefore, the system provides a supervised approach that clinicians directly control to ensure personalized patient care in radiotherapy while maintaining consistent quality.

\section{Conclusion}
\label{convlu}
In this paper, we have proposed a novel deep learning model, DoseGNN, to predict the clinical delivered doses from images for DVH prediction in automatic radiation treatment planning. DoseGNN utilizes an image conversion pipeline to provide structured representations of the geometries of OAR and PTV. Then an image-dose graph is formulated to combine the strength of LLMs in understanding doctors’ prescriptions/instructions and the effectiveness of advanced image models (like Swin Transformer) in image representation learning. Experiments demonstrate the superiority of proposed DoseGNN in accurate DVH prediction.

\newpage

\bibliography{iclr2024_conference}
\bibliographystyle{iclr2024_conference}

\clearpage
\appendix
\section{Backgrounds}

\subsection{Graph Neural Networks}

Graphs are relational data which model a set of objects (nodes) and their relationships (edges). Graph Neural Networks (GNNs) have been the dominant architectures for machine learning over graphs. In the context, GNNs have achieved strong performance in various graph learning applications \cite{you2018graphrnn, you2020handling,zitnik2017predicting, dong2022cktgnn, Li-nc}. 

As some of the earliest works, spectral GNNs ~\cite{bruna2013spectral, defferrard2016convolutional} generalize the convolution operation to graphs and define graph convolution operations from the perspective of graph signaling processing. However, spectral GNNs usually rely on the spectral decomposition (i.e. eigendecomposition) of graph Laplacian and operate on the graph spectra. Thus, spectral GNNs are computationally complex and it is hard to apply them to graphs with varying characteristics or large-scale graphs. To tackle these challenges, spatial GNNs define each node's receptive field based on the node's spatial relations and implement localized information aggregation operation within this field. Though various spatial GNNs have been proposed, such as diffusion GNN ~\cite{atwood2015diffusion}, capsule GNN ~\cite{verma2018graph}, and attention-based GNN  ~\cite{velivckovic2017graph}, most GNNs (i.e. message passing GNNs) follow the message passing framework~\cite{gilmer2017neural}. Overall, message passing GNNs are composed of some message passing layers and a readout layer. 

\paragraph{Message passing layer.} Message passing layers iteratively pass messages between each node and its neighbors to extract a node representation that encodes the local substructure. In large part due to the simplicity and scalability, the scheme shows impressive graph representation learning ability. Let $h^{t}_{v}$ denote the node features of $v$ in layer $t$, the massage passing scheme is given by:
\begin{align}
    a^{t+1}_{v} = \mathcal{A} (\{h^{t}_{u} | (u,v) \in E\}) \ \ \ 
    h^{t+1}_{v} = \mathcal{U} (h^{t}_{v}, a^{t+1}_{v}) 
\end{align}
Here, $\mathcal{A}$ is an aggregation function on the multiset of node features in $\mathcal{N}(v)$, where $\mathcal{N}(v) = \{u \in V | (u,v) \in E\}$ denotes the set of neighboring nodes of $v$, and $\mathcal{U}$ is an update function. 

\paragraph{Readout layer} After $T$ iterations/layers of the message passing process, the final node representations $\{h^{T}_{v} | v \in V \}$ are summarized into a vector $z$ 
through a readout function (e.g. mean, sum, max) as the representation of the whole graph:
\begin{align}
    z = \textit{Readout}(\{h^{T}_{v} | v \in V\})
\end{align}

\subsection{Transformers, Vision Transformers, and Large Language Models}

\paragraph{Transformers} Transformer model solves the language modeling problem \citep{dai2019transformer,al2019character,devlin2019bert,lewis2020bart} using self-attention mechanism, and improves the performance over RNN-based or convolution-based deep learning models in both accuracy and efficiency. The Transformer encoder consists of a stack of Transformer encoder layers, where each layer is composed of two sub-networks: a (multi-head) self-attention network and a feed-forward network (FFN). 

let $H = (h^{T}_{1}, h^{T}_{2}, ... h^{T}_{n})$ be the input to a Transformer encoder layer. In the self-attention network, the attention mechanism takes $H$ as input and implements different linear projections to get the query matrix $K$, key matrix $K$ and value matrix $V$, Then the attention matrix $A$ is computed as following to measure the similarities, which is then used to update the representation in parallel . 
\begin{small}
\begin{align}
    A = \frac{QK^{T}}{\sqrt{d_{k}}}, \ \ \ Z = \textit{softmax}(A)V
\end{align}
\end{small}
After the self-attention network, the feed-forward network consists of two linear transformations with a Rectified Linear Unit (ReLU) activation in between to generate the output. i.e. $O = \textit{FFN}(Z)$. The FFN is composed of a standard Dropout Layer $\to$ Layer Norm $\to$ FC (fully connected) Layer
$\to$ Activation Layer $\to$ Dropout Layer $\to$ FC Layer $\to$ LayerNorm sequence, with residual connections
from $Z$ to after the first dropout, and from before the first FC layer to after the
dropout immediately following the second FC layer. 

\paragraph{Vision Transformers} Vision Transformers employs a Transformer-like architecture over patches of the image.Transformer has been widely used in image processing, and has achieved the state-of-art performance on various image learning tasks, including object detection \citep{carion2020end} and image recognition \citep{dosovitskiy2020image}. Swin Transformer ~\cite{liu2021swin} takes the hierarchical architecture and serves as a general-purpose backbone for computer vision tasks.

\paragraph{Large Language Models (LLMs)} LLMs are also constructed upon the Transformer architecture. Due to the superior semantic comprehension capability, LLMs exhibit massive context-aware knowledge and have revolutionized workflows to handle text data. Some LLMs (e.g. Sentence-BERT ~\cite{reimers2019sentence}, Deberta~\cite{he2020deberta} LLaMA~\cite{touvron2302llama}) enable users to manipulate model's parameters and embeddings, while others (e.g. ChatGPT ~\cite{koubaa2023gpt,brown2020language}) only provide restricted access to APIs and interfaces. 

\end{document}